\documentclass{article}

\PassOptionsToPackage{numbers, compress}{natbib}

\usepackage[preprint]{neurips_2025}

\usepackage[utf8]{inputenc} 
\usepackage[T1]{fontenc}    
\usepackage{hyperref}       
\usepackage{url}            
\usepackage{bm}
\usepackage{booktabs}       
\usepackage{amsfonts}       
\usepackage{nicefrac}       
\usepackage{microtype}      
\usepackage{xcolor}         
\usepackage{enumitem}
\usepackage{adjustbox}
\usepackage{amsmath}
\usepackage{amssymb}
\usepackage{amsthm}
\usepackage{cleveref}
\usepackage{wrapfig}
\usepackage{stackengine}
\usepackage{comment}
\usepackage{mathtools}
\usepackage{xcolor}

\usepackage{graphicx}
\usepackage{caption}
\usepackage{subcaption}

\hypersetup{
    colorlinks,
    linkcolor={red!50!black},
    citecolor={blue!50!black},
    urlcolor={blue!80!black}
}

\usepackage{bm, dsfont}

\let\hat\widehat

\makeatletter
\newcommand{\ve}{\@ifnextchar\bgroup{\velong}{{\bm{e}}}}
\newcommand{\velong}[1]{{\bm{#1}}}
\makeatother

\usepackage{tikz}

\theoremstyle{plain}
\newtheorem{theorem}{Theorem}

\theoremstyle{remark}

\definecolor{LightSteelBlue1}{RGB}{202,225,255}
\definecolor{LightPink}{RGB}{245,191,210}
\definecolor{Moccasin}{RGB}{255, 228, 181}

\title{Look Within or Look Beyond? A Theoretical Comparison Between Parameter-Efficient and Full Fine-Tuning}

\author{
  Yongkang Liu \textsuperscript{\normalfont 1}$^{\ast}$ \quad
  Xingle Xu \textsuperscript{\normalfont 1}\thanks{Equal contribution} \quad
  Ercong Nie \textsuperscript{\normalfont 2} \quad
  Zijing Wang \textsuperscript{\normalfont 1} \quad
  Shi Feng \textsuperscript{\normalfont 1} \quad
  \\
  {\bf Daling Wang} \textsuperscript{\normalfont 1}\thanks{Corresponding author} \quad
  {\bf Qian Li} \textsuperscript{\normalfont 3} \quad
  {\bf Hinrich Schütze} \textsuperscript{\normalfont 2} \quad
  \\
  \textsuperscript{1}Northeastern University, China \\
  \textsuperscript{2}CIS, LMU Munich; MCML, Germany \\
  \textsuperscript{3}Shandong University, China \\
}

\begin{document}

\maketitle

\begin{abstract}

Parameter-Efficient Fine-Tuning (PEFT) methods achieve performance comparable to Full Fine-Tuning (FFT) while requiring significantly fewer computing resources, making it the go-to choice for researchers. We find that although PEFT can achieve competitive results on some benchmarks, its performance falls short of FFT in complex tasks, such as reasoning and instruction-based fine-tuning. In this paper, we compare the characteristics of PEFT and FFT in terms of representational capacity and robustness based on optimization theory. We theoretically demonstrate that PEFT is a strict subset of FFT. By providing theoretical upper bounds for PEFT, we show that the limited parameter space constrains the model's representational ability, making it more susceptible to perturbations. Experiments on 15 datasets encompassing classification, generation, reasoning, instruction fine-tuning tasks and 11 adversarial test sets validate our theories. We hope that these results spark further research beyond
the realms of well established PEFT. The source code is in the anonymous Github repository\footnote{https://github.com/misonsky/PEFTEval}.

\end{abstract}

\section{Introduction}
Full Parameter Fine-Tuning (FFT) has long been the preferred choice for fine-tuning language models due to its excellent performance~\cite{NIPS20173f5ee243,LewisLGGMLSZ20,LiuGGLEGLZ20,liu2024hift}. However, with the advent of large language models (LLMs)~\cite{zeng2022glm,vicuna2023,lin2024mala}, the GPU memory requirements for FFT have become increasingly demanding.

To reduce the memory requirements for fine-tuning language models, several alternative Parameter-Efficient Fine-Tuning (PEFT) methods to FFT have been proposed. PEFT-based approaches, such as addition-based, selection-based, and reparametrization-based methods, achieve memory savings on GPUs by reducing the number of trainable parameters~\cite{lialin2023scaling}. The addition-based tuning methods (e.g., Prefix-Tuning~\cite{li2021prefix}, AttentionFusion~\cite{cao2022attention}) reduce the number of trainable parameters by updating only newly added parameters while freezing the weights of language models. Selection-based tuning methods (e.g., BitFit~\cite{zaken2022bitfit}, LT-SFT~\cite{ansell2022composable}, FAR~\cite{vucetic2022efficient}) update only a subset of model parameters. Reparametrization-based tuning methods (e.g., LoRA~\cite{hu2021lora}, KronA~\cite{edalati2022krona}, S4-model~\cite{chen2023parameter}) reduce the number of trainable parameters by employing low-rank approximations instead of full weight matrices in learned weight updates.

These PEFT methods mitigate the high costs by updating only a small fraction of the weights~\cite{han2024parameter}. However, optimization theory suggests that models with high capacity, including overparameterized models, often demonstrate stronger representational and robustness capabilities~\cite{kohavi1996bias,dar2021farewell,neal2018modern}. This implies that fine-tuning models by reducing trainable parameters may compromise their capacity and robustness. Existing studies have shown that PEFT-based tuning performs less effectively than FFT on complex tasks~\cite{Finetuning2023,FinetuningLLMs2023,ComprehensiveFinetuningLLMs2023}. In light of the growing popularity of PEFT fine-tuning, investigating its reliability is of vital importance to the artificial intelligence community.

Preliminary research find that the incremental weight distribution for PEFT is steeper,
while the distribution of FFT in a comparatively flatter curve. 
According to the sharpness-flatness theory~\cite{Foret2020SharpnessAwareMF}, models with flatter parameter distributions exhibit greater robustness, whereas those with steeper distributions are more sensitive to perturbations. Simultaneously, FFT’s abundant degrees of freedom confer stronger representation ability than the limited-capacity PEFT. We prove that PEFT is a subset of FFT, and the upper bound of PEFT reveals that the constrained parameter space limits the model's representational ability and makes it more susceptible to perturbations.
Experimental results on 15 datasets, including classification, generation, reasoning, instruction fine-tuning tasks, and 11 adversarial test sets, validate our theories. 
Our main contributions and findings are summarized as follows:

\begin{itemize}[leftmargin=*]
  \item We prove that PEFT is a strict subset of FFT, which is a low-dimensional, measure-zero manifold inside the FFT.
  \item We prove that PEFT’s representational capacity is bounded by a well-defined upper bound, the number of fine-tuning parameters in PEFT directly constrains the maximum extent of model adaptation, which may constrain the model’s capacity for learning novel knowledge.
  \item We prove that PEFT’s limited parameter space inherently caps its data-driven gains, making its marginal improvement from extra samples much lower than FFT’s.
  \item We demonstrate that PEFT’s incremental weight distribution is both steeper and more sharply peaked, and we theoretically prove it is more susceptible to perturbations than FFT.
\end{itemize}

\section{Birds-eye View of Fine-Tuning}

\paragraph{Full Parameter Fine-Tuning} refers to updating all parameters in full rank. In full-rank optimization, local critical points are typically saddle points rather than valleys~\cite{dauphin2014identifying}. Saddle points are characterized by having exits in multiple dimensions, which means that models are more likely to escape them and converge towards the global optimal solution~\cite{dauphin2014identifying}. Generally, once the dataset and network architecture are specified, the optimization space of the model remains fixed. Subsequent parameter updates by the optimizer explore the fine details within this frozen space~\cite{li2018measuring}.

We represent the foundational model as a function $f(x,\theta)$, where $\theta \in \Theta = \mathbb{R}^d$, where $d$ denotes the spatial dimension. Given a data $D$, the purpose of fine-tuning is to identify a parameter $\theta^{\prime}$ satisfying the following equation:
\begin{equation}
\label{equation:1}
\theta^{\prime} = \text{arg min}_{\theta \in \Theta} \mathbb{E}_{(x,y)\thicksim D}[\mathcal{L}(f(x;\theta),y)]
\end{equation}
where $\mathcal{L}$ is the loss function. The model of FFT can be defined as a set of functions:
\begin{equation}
\label{equation:2}
\mathcal{F}_{full}={f(x;\theta):\theta \in \Theta}
\end{equation}
Its optimization space is $\mathbb{R}^d$.
\paragraph{Parameter-Efficient Fine-Tuning} refers to the process of fine-tuning only a subset of parameters $\Phi$ (newly introduced or part of the base model) to adapt to downstream tasks. While reducing the number of trainable parameters during fine-tuning can decrease memory usage, it also limits the model's capacity for representation. Compared to FFT, PEFT updates occur in a relatively lower-dimensional space. This implies that once the model encounters a saddle point during fine-tuning, it may be more challenging to escape. However, this does not necessarily indicate that PEFT is more prone to local optima, as high-dimensional space may harbor more saddle points than low-dimensional space. 

The model of PEFT can be expressed as $f(x;\theta_0;\Phi)$, where $\theta_0$ is frozen and represents the initial state of the foundational model. For any PEFT method, assume that the spatial dimension of incremental parameter $\Phi$ is $k$, where $k \ll d$. The model of PEFT can be defined as a set of functions:
\begin{equation}
\label{equation:3}
\mathcal{F}_{peft}={f(x;\theta_0;\Phi):\Phi \in \mathbb{R}^k \subset \Theta}.
\end{equation}
Fundamentally, PEFT can be viewed as a parameter reparameterization mechanism applied to the model, namely:
\begin{equation}
\label{equation:4}
f(x;\theta_0;\Phi)=f(x;\theta_0+ g(\Phi)).
\end{equation}
where $g:\mathbb{R}^k \rightarrow \mathbb{R}^d$ is a mapping from a low-dimensional space to a high-dimensional space, typically with a linear or sparse structure (e.g., LoRA~\cite{hu2021lora}, Adapter~\cite{HuWLXLB0PL23}). 

\section{Theory}
Our theoretical analysis highlights why FFT is more reliable than PEFT in terms of model representation capacity and robustness, and some classical results have shown the limited parameter space in PEFT constrains the model's performance\cite{huang2025hira,he2024parameter,qiao2024learn}. First, we show that the fine-tuning space of PEFT is an embedded submanifold of the parameter space of FFT, and prove that the PEFT fine-tuning subspace is a strict subset of that of FFT (\Cref{theorem:1}). At the same time, we presents upper bound on the representation capacity of PEFT fine-tuning, which implies that the number of fine-tuning parameters in PEFT directly constrains the maximum extent of model adaptation (\Cref{theorem:2}). Then, we demonstrate that there exists an optimal number of parameters for PEFT fine-tuning, beyond which increasing the number of tunable parameters yields diminishing returns (\Cref{theorem:3}). From a robustness perspective, our theory indicates that PEFT is more sensitive to perturbations compared to FFT (\Cref{theorem:4}). From a data perspective, we show that the limited parameter space of PEFT constrains its ability to benefit from additional data, resulting in smaller marginal gains from sample increases compared to FFT (\Cref{theorem:5}).

\begin{theorem}
\label{theorem:1}
\emph{(Subset PEFT of FFT) }
According to equation~\ref{equation:4}, we define $\theta_{\Phi} := \theta_0 + g(\Phi) \in \mathbb{R}^d$, where $g$ is a non-surjective function (proof in the \Cref{proof:1}). The conclusions can be drawn:
\begin{equation}
f(x; \theta_0; \Phi) = f(x; \theta_\Phi) \Rightarrow f(x; \theta_0; \Phi) \subset \mathcal{F}_{full}.
\end{equation}
That is to say, $\forall \Phi \in \mathbb{R}^k, \; \exists \theta_{\Phi} \in \mathbb{R}^d \; \text{such that} \; f(x; \theta_0; \Phi) = f(x; \theta_{\Phi}) \subset \mathcal{F}_{full}$

\end{theorem}

\begin{theorem}
\label{theorem:2} 
\emph{(PEFT Capacity Upper Bound) }
Assume that the upper bounds of the weight norm and activation Lipschitz constant of each layer are $M$ and $L$ respectively, that is, $\|W_{0,k}\| \leq M, \quad L_k \leq L, \quad \forall k$. At the same time, let $\alpha_k \triangleq L \|\Phi_k\|$, for input $x \in D$, we have (proof in the \Cref{proof:2}):
\begin{equation}
\|f(x) - f_0(x)\| \leq \sum_{k=1}^{N} M^{N-k} L^{N-k} \alpha_k \|x\|.
\end{equation}
This theorem demonstrates that the number of fine-tuning parameters in PEFT directly constrains the maximum extent of model adaptation, with this relationship quantified through the term 
$\alpha_k$. We observe that the $M^{N-k} L^{N-k}$ term in the formula indicates that PEFT parameters across different layers have unequal effects on the final output. Specifically, parameters closer to the output layer (i.e., with larger $k$ values) have a greater impact, as the exponent $N - k$ becomes smaller.
\end{theorem}
\begin{theorem}
\label{theorem:3}
\emph{(Rule of Diminishing Marginal Benefit) }
In PEFT fine-tuning, there exists a critical parameter $\Phi_c$ such that when $Rank(\Phi) \geq r_c$ ($Rank(\cdot)$ is the rank function.), further increasing the parameter amount of $\Phi$ does not lead to significant improvements in the performance of the optimal solution to the optimization problem.
Assume $N$ is the number of training samples, $r \ge r_c$, the following conclusions can be drawn (proof in the \Cref{proof:3}):
\begin{equation}
\mathcal{L}_{\mathcal{D}_{test}}(W + \Delta W_r) \leq \mathcal{L}_{\mathcal{D}_{test}}(W + \Delta W^*) + L\sqrt{\epsilon} + O(\sqrt{\frac{r}{N}}).
\end{equation}
where $\Delta W^*$ is ideal weight update. As $r$ continues to increase, the first term remains unchanged, the second term approaches zero, but the third term (generalization error) increases, leading to a plateau or even a decline in overall performance.
\end{theorem}
Together, experiments show that once the critical number of parameters is reached, the model performance remains stable even if the parameters continue to increase.

\begin{theorem}
\label{theorem:4}
\emph{(PEFT More Sensitive to Disturbances) }
Assume the disturbance factor is $\epsilon$, the loss fluctuations induced by the perturbation factor $\epsilon$ in PEFT and FFT fine-tuning are as follows (proof in the \Cref{proof:4}):
\begin{equation}
\begin{aligned}
\Delta\mathcal{L}_{\text{full}} &\approx \frac{1}{2}\Delta\theta^T H\Delta\theta, \\
\Delta \mathcal{L}_{peft} \approx \frac{1}{2}&\epsilon_{\parallel}^TH_{\Phi}^{+}\epsilon_{\parallel} + \epsilon_{\perp}^T \nabla_\theta\mathcal{L}.
\end{aligned}
\end{equation}
where $H$ is the Hessian Matrix, $H_{\Phi}^{+}$ is the pseudo-inverse of $P^THP$, $\epsilon_{\parallel}$ is the projection of the perturbation in the PEFT subspace, $\epsilon_{\perp}$ is the component orthogonal to this subspace. We have $\Delta\mathcal{L}_{\text{peft}} - \Delta\mathcal{L}_{\text{full}}> 0$.
\end{theorem}
Results across multiple adversarial datasets show that PEFT is more sensitive to perturbations than FFT.

\begin{theorem}
\label{theorem:5}
\emph{(Marginal Benefit of Examples) } The limited parameter space of PEFT constrains its data-driven gains, resulting in smaller benefits from increased data compared to FFT.

According to Equations~\ref{equation:1} to ~\ref{equation:4}, the optimal solutions of PEFT and FFT in Empirical Risk Minimization (ERM) are:
\begin{equation}
\begin{aligned}
\hat{\theta}_{N} &= \arg\min_{\theta\in\mathbb{R}^d} \frac{1}{N} \sum_{i=1}^{N} \ell(f_\theta(x_i), y_i), \\
\hat{\Phi}_{N} &= \arg\min_{\Phi\in\mathbb{R}^k} \frac{1}{N} \sum_{i=1}^{N} \ell(f_{\theta_0+g(\Phi)}(x_i), y_i).
\end{aligned}
\end{equation}
where $N$ represents the number of training sets. The expected risk of ERM can be decomposed into:
\begin{equation}
\underbrace{\mathcal{L}(\hat{\theta}_N)}_{\text{expected error}} - \underbrace{\mathcal{L}^*_{\mathcal{H}}}_{\text{approximation error}} = \underbrace{\mathcal{L}(\hat{\theta}_N) - \mathcal{L}_N(\hat{\theta}_N)}_{\text{generalization error}} + \underbrace{\mathcal{L}_N(\hat{\theta}_N) - \mathcal{L}^*_{\mathcal{H}}}_{\text{empirical optimality achievement error}}.
\end{equation}
where $\mathcal{L}(\hat{\theta}_N)$ represents the loss of the model's true distribution, $\mathcal{L}^*_{\mathcal{H}}$ represents the optimal loss of the model in the hypothesis space $\mathcal{H}$ and $\mathcal{L}_N(\hat{\theta}_N)$ represents the loss on the training sample. We make the assumption that the model has already been trained for many steps on the fine-tuning objective, which means $(\mathcal{L}_N(\hat{\theta}_N) - \mathcal{L}^*_{\mathcal{H}}) \to 0$, then:
\begin{equation}
\underbrace{\mathcal{L}(\hat{\theta}_N)}_{\text{expected error}} = \underbrace{\mathcal{L}(\hat{\theta}_N) - \mathcal{L}_N(\hat{\theta}_N)}_{\text{generalization error}} + \underbrace{\mathcal{L}^*_{\mathcal{H}}}_{\text{approximation error}}.
\end{equation}
Rademacher complexity~\cite{blum2007machine} theory gives us:
\begin{equation}
\mathbb{E} \left[ \mathcal{L}(\hat{\theta}_N) - \mathcal{L}_N (\hat{\theta}_N) \right] = O\left(\sqrt{\frac{d}{N}}\right).
\end{equation}
For FFT, the approximation error and generalization error are:
\begin{equation}
	\begin{aligned}
		A_{full} = \mathcal{L}_{\Theta}^*, \quad E_{full}(N) = O\left(\sqrt(d/N)\right).
	\end{aligned}
\end{equation}
Similarly, for PEFT we have:
\begin{equation}
	\begin{aligned}
		A_{peft} = \mathcal{L}_{\Phi}^*, \quad E_{peft}(N) = O(\sqrt{\frac{k}{N}}).
	\end{aligned}
\end{equation}
Their expected risks are:
\begin{equation}
R_{full} = A_{full} + E_{full}(N), \quad R_{peft} = A_{peft} + E_{peft}(N).
\end{equation}
The decrease in expected risk resulting from increasing the sample size from $N$ to $N+1$ is characterized by:
\begin{equation}
\begin{aligned}
\Delta R_{full}(N) = R_{full}(N)-R_{full}(N+1), \quad \Delta R_{peft}(N) = R_{peft}(N)-R_{peft}(N+1).
\end{aligned}
\end{equation}
Ignoring constant factors and taking the limit as $N \to N+1$, we obtain:
\begin{equation}
\begin{aligned}
\frac{d}{dN}E_{\text{full}}(N) &\approx -\frac{1}{2}\sqrt{\frac{d}{N}}N^{-3/2} = -O(d\,N^{-3/2}), \quad
\frac{d}{dN}E_{\text{peft}}(N) &\approx -O(k\,N^{-3/2}).
\end{aligned}
\end{equation}
The approximation errors $A_{full}$ and $A_{peft}$ are independent of $N$, so we have:
\begin{equation}
\Delta R_{\text{full}}(N) \approx -\frac{d}{dN}E_{\text{full}}(N) = O(d\,N^{-3/2}), \quad \Delta R_{\text{peft}}(N) = O(k\,N^{-3/2}).
\end{equation}
Since $k \ll d$, we can get:
\begin{equation}
\frac{\Delta R_{\text{full}}(N)}{\Delta R_{\text{peft}}(N)} = O\left(\frac{k}{d}\right) \ll 1.
\end{equation}
That is, for each additional sample, the risk reduction achieved by PEFT is $k/d$ times that of FFT, which is significantly smaller.
\end{theorem}

\begin{wraptable}{r}{7cm}
\caption{Performance comparison of different fine-tuning methods based on LLaMA2 family. All reported numbers are averaged metrics (standard deviation).}
\label{tab:llama-math}
\centering
\begin{adjustbox}{max width=0.45\textwidth, center}
\includegraphics[width=\textwidth]{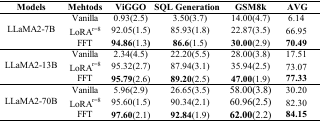}
\end{adjustbox}
\end{wraptable}

\begin{table*}[ht]
  \caption{Comparison of instruction fine-tuning on mt-bench with different fine-tuning methods.}
  \label{tab:mt-bench}
	\begin{adjustbox}{max width=\textwidth, center}
		\includegraphics[width=\textwidth]{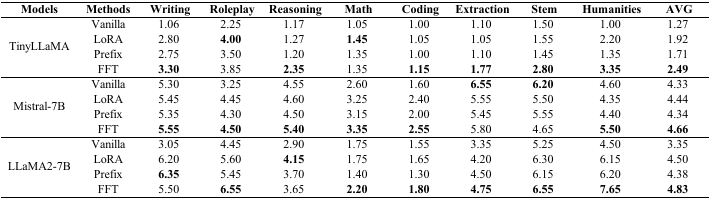}
	\end{adjustbox}
\end{table*}

\begin{table*}[ht]
\caption{Performance comparison of different fine-tuning methods based on RoBERTa$_\text{base}$ and RoBERTa$_\text{large}$ on AdvGLUE dataset. All reported numbers are averaged metrics (standard deviation).}
  \label{tab:adv-glue}
	\begin{adjustbox}{max width=\textwidth, center}
		\includegraphics[width=\textwidth]{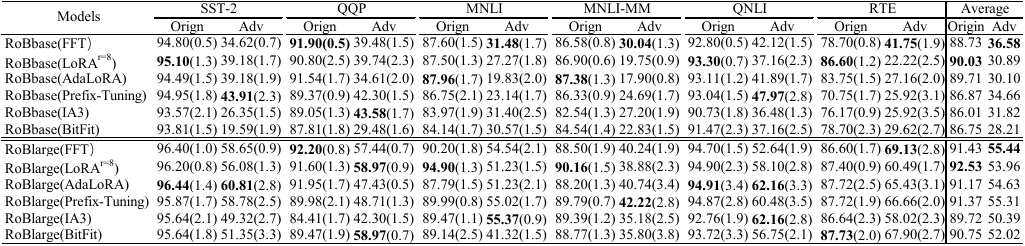}
	\end{adjustbox}
\end{table*}
\section{Experiments}
\Cref{app:base} reports all used language models and fine-tuning methods. All experiments below use datasets detailed in \Cref{app:datasets}. 
All fine-tuning with backpropagation experiments follow convention and use Adam, the detailed settings including hyperparameters are reported in \Cref{app:implementation_details}. Experiments below use prompts detailed in \Cref{app:prompt}. We also discuss the limitations of this paper (\Cref{sec:limitations}). 

We conduct comprehensive experiments on both medium-sized masked language models (RoBERTa-base, RoBERTa-large~\cite{abs190711692}, and TinyLLaMA~\cite{Llama2}) and large autoregressive language models (Mistral-7B~\cite{jiang2023mistral}, OPT-13B~\cite{abs220501068}, LLaMA2-7B, LLaMA2-13B and LLaMA2-70B~\cite{Llama2}). First, we show that FFT has performance advantages over PEFT in complex tasks such as math reasoning, SQL generation, instruction fine-tuning, and functional representation. Moreover, FFT performs more robustly on adversarial datasets than PEFT. Further experiments demonstrate that the marginal benefit of FFT for increasing data and parameters is greater than that of PEFT.

\begin{table*}[ht]
\caption{Performance comparison of different fine-tuning methods based on RoBERTa$_\text{base}$ and RoBERTa$_\text{large}$ on Adversarial SQuAD dataset.}
  \label{tab:adv-squad}
	\begin{adjustbox}{max width=0.98\textwidth, center}
		\includegraphics[width=\textwidth]{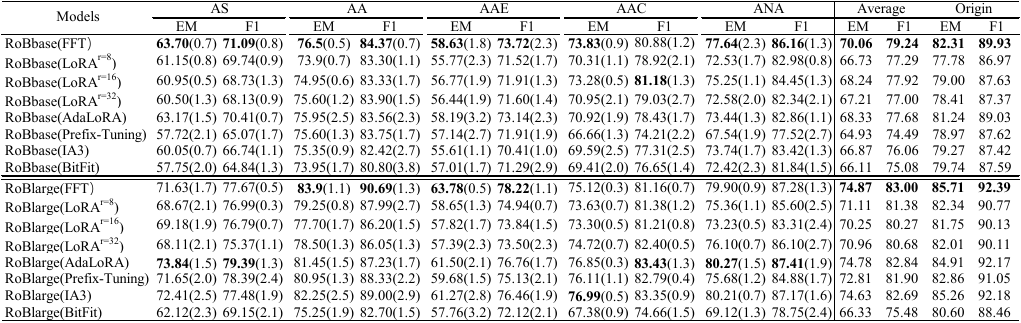}
	\end{adjustbox}
\end{table*}

\subsection{Performance Comparison Between FFT And PEFT}
We compare the performance of different fine-tuning methods on multiple datasets across various tasks. Table~\ref{tab:llama-math} presents the performance comparison of LLaMA2-7B, LLaMA2-13B, and LLaMA2-70B using different fine-tuning methods. Across all LLaMA2 models, standard FFT consistently achieves a performance advantage over LoRA fine-tuning. Specifically, FFT-based LLaMA2-7B shows a 2.81\% higher performance than LoRA on ViGGO, 0.67\% higher on SQL Generation, and 7.31\% higher on GSM8k. Similar trends are observed in LLaMA2-13B and LLaMA2-70B. On average, FFT-based LLaMA2-7B outperforms LoRA by 3.54\%, LLaMA2-13B outperforms by 4.26\% and LLaMA2-70B outperforms by 1.85\%. As model size increases, FFT consistently maintains its performance advantage over LoRA.

Table~\ref{tab:mt-bench} presents the instruction fine-tuning results of TinyLLaMA, Mistral-7B, and LLaMA2-7B using different fine-tuning methods. FFT-based TinyLLaMA shows performance advantages over PEFT methods, including LoRA, in 6 out of 8 dimensions. Similar trends are observed for LLaMA2-7B and Mistral-7B. Compared to LoRA, FFT-based TinyLLaMA achieves an average performance improvement of 0.57, Mistral-7B by 0.22, and LLaMA2-7B by 0.33. FFT demonstrates a consistent performance advantage over PEFT in the instruction fine-tuning task. Assuming our fine-tuning is effective, according to \Cref{theorem:2}, we have: 
\begin{equation}
f_0(x) \leq f(x) \leq \sum_{k=1}^{N} M^{N-k} L^{N-k} \alpha_k \|x\| + f_0(x).
\end{equation}
Since the values of $N$, $M$, and $\alpha_k$ in FFT are higher than those in PEFT, FFT demonstrates stronger upper bound that PEFT. When tackling complex tasks, models often need to learn intricate features to make accurate decisions. The FFT offers a more powerful representation space, allowing the model to better capture and utilize these features.

We also observe that LoRA achieves performance advantages over FFT on the original GLUE dataset as shown in Table~\ref{tab:adv-glue}, while LoRA performs inferiorly to FFT on SQuAD as indicated in Table~\ref{tab:adv-squad}. It's important to emphasize that PEFT methods, including LoRA, do not consistently work well, particularly on complex tasks. 
Simple tasks with straightforward formats and clear label characteristics may not necessitate a large number of parameters to effectively learn data features; A reduced parameter count can still capture sufficient label-related information. This is where PEFT demonstrates efficacy in certain benchmarks.
\begin{figure}[t]
\centering
\begin{adjustbox}{max width=0.95\textwidth}
\includegraphics[width=\linewidth]{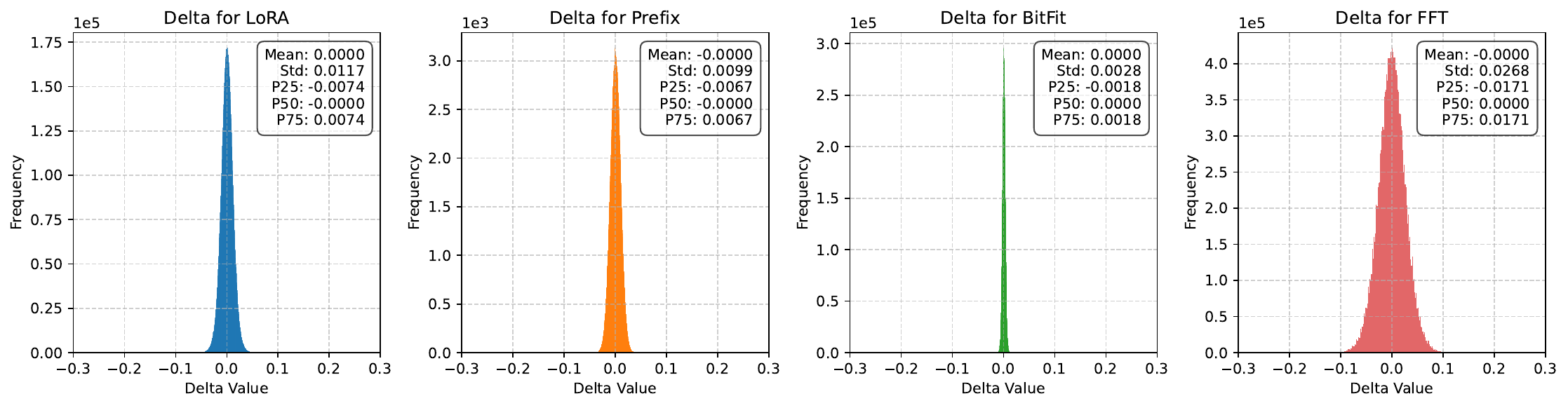}
\end{adjustbox}
\caption{Incremental parameter distributions for different fine-tuning methods. For Prefix, the incremental parameters are prompt embeddings. For LoRA, BitFit, and FFT are the query of the last layer. The same phenomenon can be observed in other layers for LoRA, BitFit, and FFT. The base model is LLaMA2-7B. The task is instruction tuning on Alpaca.}
\label{fig:para-distributions}
\end{figure}

\begin{table*}[ht]
\caption{Performance comparison of different fine-tuning methods based on OPT-13B (with 1000 examples) on Adversarial SQuAD dataset.}
  \label{tab:opt-squad}
	\begin{adjustbox}{max width=\textwidth, center}
		\includegraphics[width=\textwidth]{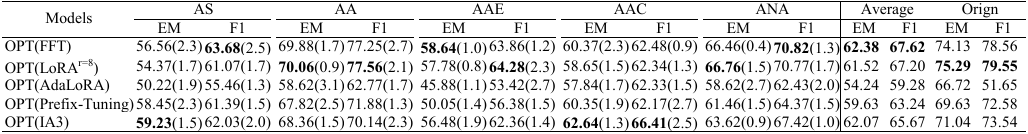}
	\end{adjustbox}
\end{table*}
\paragraph{Parameter Distribution Comparison Between FFT and PEFT}
\label{sub:parameter_distribution}
To further explore the differences between FFT and PEFT, Figure~\ref{fig:para-distributions} shows the parameter distribution of various fine-tuning methods. We can observe that the incremental parameters of PEFT (including LoRA, prefix-tuning, BitFit) exhibit a steeper distribution around zero, with the Standard Deviation value being close to zero. This suggests that the adjustments made by these incremental parameters have a minimal impact on the underlying model. The incremental parameters of FFT display the widest distribution compared to PEFT, indicating a greater capacity for parameter adjustment. This pattern suggests that PEFT methods primarily adapt the model to follow instruction formats rather than incorporating substantial new knowledge. The minimal parameter changes indicate these techniques are effectively repurposing existing knowledge in the base model rather than learning new information during fine-tuning.

\begin{wrapfigure}[]{r}{0.56\textwidth}
	\centering
	\begin{adjustbox}{max width=1\textwidth}
		\includegraphics[width=\linewidth]{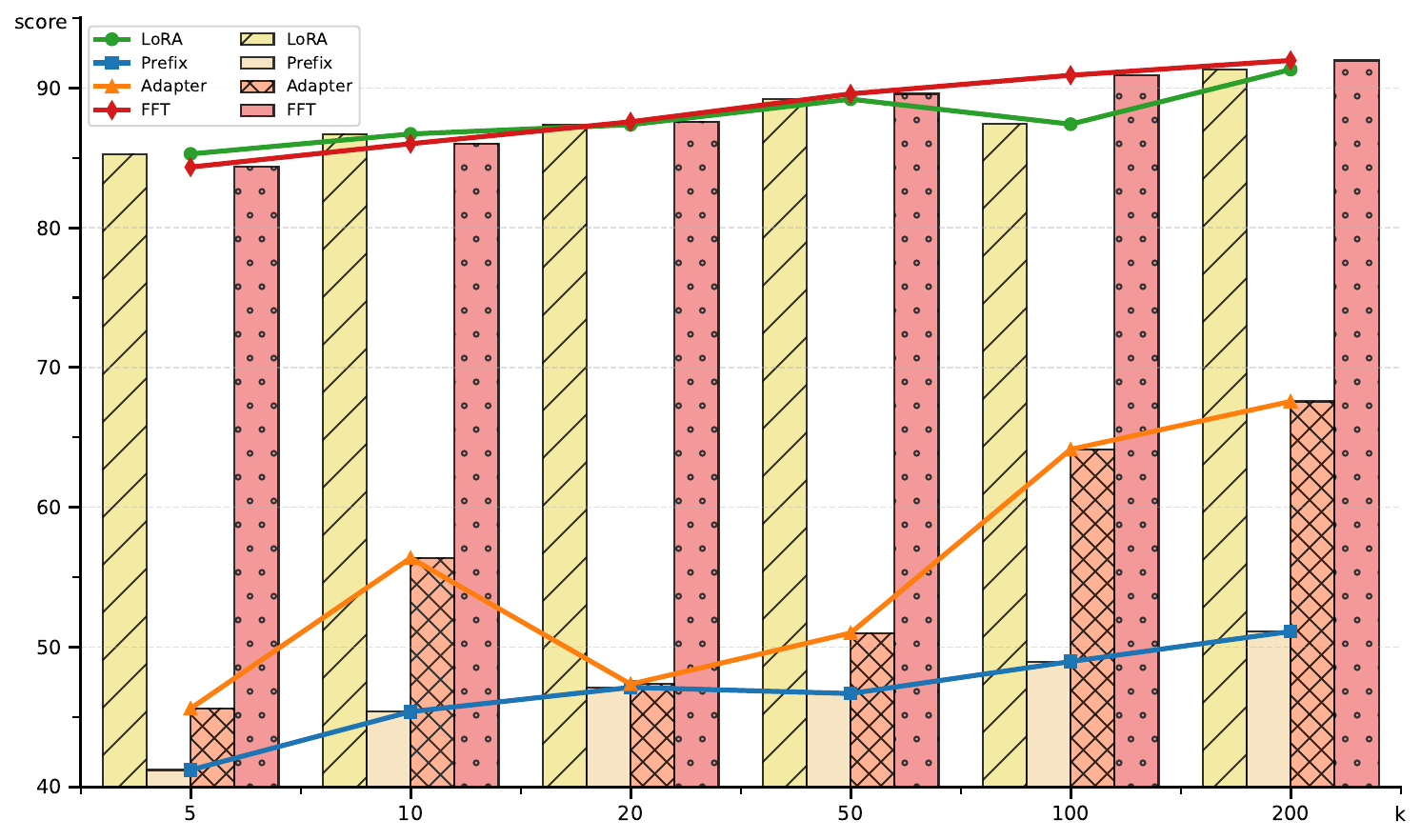}
	\end{adjustbox}
	\caption{The impact of training samples on the performance of different fine-tuning methods. $k$ denotes the number of training examples per class. The score represents the average performance of different fine-tuning methods on all test set. The base model is LLaMA2-7B.}
	\label{fig:FFT-trend}
\end{wrapfigure}

\subsection{Robustness Comparison Between FFT And PEFT}
Tables \ref{tab:adv-glue} and \ref{tab:adv-squad} present the performance comparison of various fine-tuning methods on the adversarial test sets AdvGLUE and Adversarial SQuAD. In AdvGLUE, we observe that RoBERTa$_\text{base}$ and RoBERTa$_\text{large}$ models fine-tuned with LoRA achieve performance advantages over FFT on the original dataset. Specifically, RoBERTa$_\text{base}$ based on LoRA performs 1.3\% better than FFT, while RoBERTa$_\text{large}$ shows a 1.1\% improvement. Other PEFT methods such as AdaLoRA, Prefix, and IA3 exhibit performance levels very close to FFT. However, FFT demonstrates an overall average performance advantage over PEFT on AdvGLUE. Specifically, RoBERTa$_\text{base}$ fine-tuned with FFT outperforms LoRA by 5.69\%, and RoBERTa$_\text{large}$ outperforms LoRA by 1.48\%.
On the Adversarial SQuAD dataset, we find that standard FFT consistently outperforms other PEFT methods on both the original dataset and the adversarial test set. We observe similar phenomena on large language models. As shown in Table~\ref{tab:opt-squad}, experiments on the Adversarial SQuAD dataset using OPT-13B show that LoRA achieves the highest performance on the original test set compared to other fine-tuning methods. However, FFT performs better than LoRA on the adversarial test set. 

\Cref{theorem:4} shows that PEFT is more sensitive than FFT under the same disturbance $\epsilon$. \Cref{theorem:1} indicates that PEFT is a subset of FFT. More specifically, PEFT is an embedded submanifold of the FFT parameter space. This means when the disturbance direction is orthogonal to the PEFT space (i.e., $\epsilon_{\parallel} = 0$), the model loses the ability to extract effective features and filter out invalid ones, which significantly impacts its performance. Generally speaking, the disturbance $\epsilon$ has components in the orthogonal direction of PEFT space and in the parameter space direction. Since the task feature perception parameters are limited to the PEFT subspace, there is nothing model can do about the orthogonal perturbation features. Due to the completeness and high degree of optimization flexibility in the FFT update parameter space, the model can more effectively learn to extract task-related features and distinguish between disturbance information. Therefore, from the perspective of user reliability, we believe that FFT is more robust than PEFT. 

Relying solely on performance metrics might lead to the misconception that PEFT methods, including LoRA, are viable alternatives to FFT. However, from a robustness perspective, FFT maintains a clear advantage. This prompts us to consider whether reducing the number of trainable parameters could potentially diminish certain aspects of the model's capabilities, such as robustness, making the model more vulnerable, or its representational capacity, hindering its ability to adequately capture data features. Until these potential risks associated with PEFT are fully explored, FFT remains a robust and dependable fine-tuning method.



\begin{figure}[ht]
\centering
\includegraphics[width=0.88\textwidth]{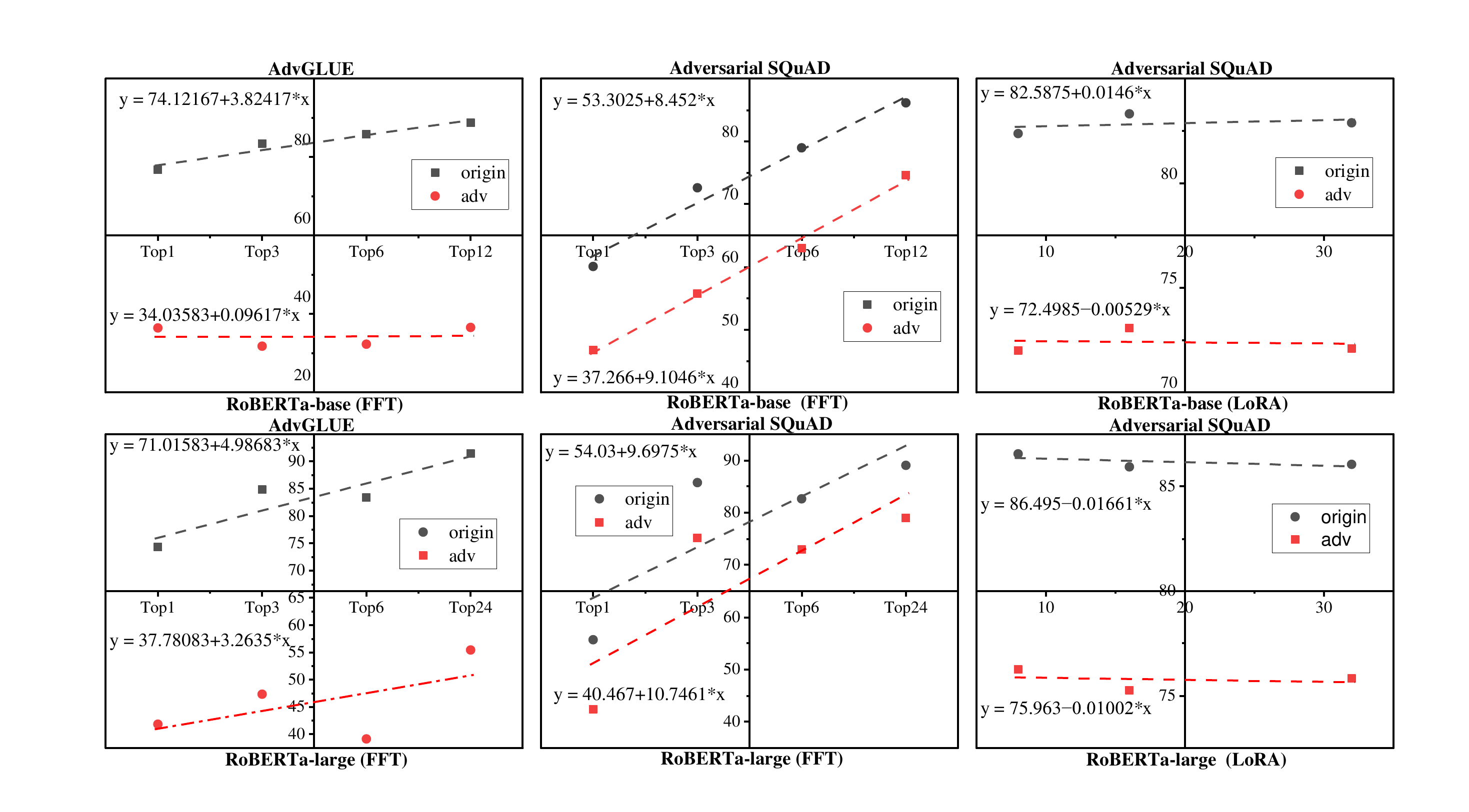}
\caption{Performance trends of FFT and LoRA fine-tuning  with different amounts of fine-tuning parameters on AdvGLUE and Adversarial SQuAD.}
\label{fig:FFT-lora-trend}
\end{figure}



\subsection{Comparison of Marginal Benefits of Data Increment}
Table~\ref{tab:llama2-few} (\Cref{sec:more_experiments}) presents the performance variations of different fine-tuning methods across various sample sizes. When  $k = 5$  and $k = 50$, FFT outperforms PEFT in two out of five tasks. When $k = 10$ and $k = 20$, FFT outperforms PEFT in only one out of five tasks. However, when $k = 100$ or $k = 200$, FFT outperforms PEFT in four out of five tasks. Figure~\ref{fig:FFT-trend} compares the average performance of different fine-tuning methods on these five benchmarks.

We observe that when $k < 20$, LoRA generally outperforms FFT. When $k = 20$, LoRA and FFT demonstrate similar performance. However, when $k > 20$, FFT begins to outperform PEFT. Based on these results, we conclude that having sufficient training data is crucial to fully utilize FFT. Under few-shot fine-tuning scenarios, the PEFT method achieves a performance advantage more readily than FFT. Increasing the training data from 5 to 200 samples results in FFT's performance increasing by 9.05\%, while LoRA's performance improves by 7.08\%. Thus, augmenting the training data significantly enhances the performance advantages of FFT.
For large language models, estimating the requisite amount of data for effective fine-tuning can be challenging~\cite{kaplan2020scaling}. According to \Cref{theorem:5}, when adding the same amount of data at the same time, the marginal benefit ratio of PEFT and FFT is $O(\frac{k}{d}) \ll 1$, where $k$ and $d$ represent the parameter dimensions of FFT of PEFT respectively.
The optimization space of FFT is complete, and its parameter updates can encompass all possible feature directions. With sufficient degrees of freedom, the model can always find unoptimized feature dimensions for new data. This ample parameter space allows for the learning of additional features. PEFT restricts optimization to a $k$-dimensional subspace, with new data being incorporated only through low-dimensional projections (such as the low-rank matrix in LoRA). This limits the capacity of the parameter space, leading to the loss of significant information. Once the low-dimensional space is saturated with features, there is no remaining parameter space to accommodate additional feature information. The marginal benefit of FFT increases significantly with growing data volume, surpassing that of PEFT.

\subsection{Comparison of Marginal Benefits of Parameter Increment}
\Cref{theorem:3} shows that PEFT has diminishing marginal benefits of parameter increments.
To investigate the impact of the number of trainable parameters on the performance of FFT and PEFT, we analyze the performance variations when fine-tuning the top-k layers and the performance changes of LoRA at different ranks $r$, where $k \in \{1, 3, 6, 12, 24\} $ and $r \in \{8, 16, 32\}$. Note that since obtaining models with the same structure but different parameter scales is challenging, we simulate the parameter changes of FFT by fine-tuning the top-k layers.

Table~\ref{tab:abla_glue}, Table~\ref{tab:abla_squad} (\Cref{sec:more_experiments}) and Figure~\ref{fig:FFT-lora-trend} report the performance changes of RoBERTa$_\text{base}$ (RoBbase) and RoBERTa$_\text{large}$ (RoBlarge) with different numbers of trainable parameters on the AdvGLUE and Adversarial SQuAD datasets, respectively.
On AdvGLUE, FFT exhibits the best average performance on both the original dataset and the adversarial test set. A similar trend is observed on Adversarial SQuAD. According to \Cref{theorem:3}:
$\mathcal{L}_{\mathcal{D}_{test}}(W + \Delta W_r) \leq \mathcal{L}_{\mathcal{D}_{test}}(W + \Delta W^*) + L\sqrt{\epsilon} + O(\sqrt{\frac{r}{N}})$
When $r < r_c$, progressively increasing $r$ can enhance the model's expressiveness, reduce both optimization and generalization errors, and improve overall performance. When $r \ge r_c$, the model's expressive power is saturated, and further increasing $r$ yields negligible improvement toward the ideal loss $\mathcal{L}_{\mathcal{D}_{test}}(W + \Delta W^*)$. As the results show, increasing the rank of LoRA shows minimal impact on the model's performance and robustness. This indicates that increasing the number of trainable parameters in LoRA fine-tuning does not effectively enhance the model's performance and robustness as it does with FFT.
\section{Related Work}
\paragraph{\textbf{Parameter-Efficient Fine-tuning}} PEFT adapts a language model (LM) to downstream tasks by freezing most of the weights and only updating a small set of internal or additional parameters, thereby minimizing resource utilization~\cite{LiuTMMHBR22,DingQYWYSHCCCYZWLZCLTLS23}. PEFT models can be classified as addition-based, selection-based, or decomposition-based methods~\cite{lialin2023scaling}.
Addition-based methods introduce and update new parameters while keeping the weights of language models frozen. Examples include LoRA~\cite{hu2021lora}, Prefix-Tuning~\cite{li2021prefix,ZhangTX00H23}, AttentionFusion~\cite{cao2022attention}, and adapters~\cite{HuWLXLB0PL23}.
Selection-based methods fine-tune a subset of the existing parameters of LMs, such as BitFit~\cite{zaken2022bitfit}, LT-SFT~\cite{ansell2022composable}, and FAR~\cite{vucetic2022efficient}. Reparametrization-based methods use low-rank decomposition to minimize the number of trainable parameters, such as PHM~\cite{karimi2021compacter}, KronA~\cite{edalati2022krona}, S4-model~\cite{chen2023parameter}, and PERU-LoRA~\cite{abs231001886}.
PEFT reduces memory usage during fine-tuning by decreasing the number of tunable parameters. However, this can potentially diminish the model's representational ability, which may negatively impact its performance and robustness.
\paragraph{Robustness Evaluation}
Using performance as the sole indicator to measure the quality of a model is inappropriate. Existing literature has shown that high-performance models, including LLMs, are vulnerable to carefully crafted adversarial examples~\cite{wang2023robustness,abs230604618,abs230212095}, which can deceive the models into producing arbitrarily incorrect answers by subtly perturbing inputs in ways imperceptible to humans. Artificial general intelligence systems built upon these vulnerable models can be easily misled, leading to significant security issues~\cite{WangXWG0GA021,wang2023robustness,zhu2023promptbench}.
Researchers have made several attempts to reveal the robustness of models~\cite{goyal2023survey}. \citet{WangXWG0GA021} propose Adversarial GLUE (AdvGLUE) to quantitatively and thoroughly explore and evaluate the vulnerabilities of language models under various types of adversarial attacks.
\citet{jia2017adversarial} and \citet{liu2020robust} provide adversarial SQuAD to evaluate the robustness of reading comprehension systems from different dimensions. In addition to adversarial datasets, integrated attack tools that include word, sentence, and multi-level adversarial attacks play a key role in assessing model robustness~\cite{ZengQZZMHZLS21}.
OpenAttack~\cite{ZengQZZMHZLS21}, TextAttack~\cite{MorrisLYGJQ20}, TextFlint~\cite{WangLGZZZYZZPWL21}, and Robustness Gym~\cite{GoelRVTBR21} integrate various ad hoc input transformations for different tasks and provide programmable APIs to dynamically test model performance. \citet{abs230212095} conduct a thorough evaluation of the robustness of ChatGPT from an adversarial and out-of-distribution~\cite{abs230604618} perspective. In summary, many works have focused on evaluating robustness. However, the robustness of the PEFT method remains unexplored.

\section*{Conclusion}
In this paper, we compare different fine-tuning methods on 15 datasets and 11 sub-adversarial datasets from the perspective of robustness and performance. We have observed that PEFT can achieve comparable or even superior performance to FFT on certain benchmarks. However, FFT consistently maintains a performance advantage over PEFT in the context of complex tasks that were evaluated. In the adversarial test, PFFT clearly demonstrates a significant performance advantage over PEFT. Therefore, we still recommend using FFT for fine-tuning models to the extent that resources allow.


\newpage

\bibliography{bibliography}
\bibliographystyle{plainnat}

\clearpage

\appendix

\section{Baselines}
\label{app:base}
\textbf{Language Models}
\begin{itemize}[leftmargin=*]
\item \textbf{RoBERTa}~\cite{abs190711692} is a transformer-based model built on BERT, designed by Facebook AI. It improves BERT by removing the Next Sentence Prediction objective and training with larger mini-batches and longer sequences. RoBERTa has set new benchmarks in NLP tasks by leveraging more data and training with different hyperparameters, making it a state-of-the-art pre-trained language model.
\item \textbf{LLaMA}~\cite{Llama2} is a family of decoder-only models by Meta, designed to be efficient for a wide range of NLP tasks. LLaMA models are trained with a focus on providing strong performance while maintaining fewer parameters, making them more accessible and computationally efficient compared to larger models like GPT-3. We use 7B, 13B, and 70B versions of LLaMA.
\item \textbf{TinyLLaMA} is a compact version of LLaMA, optimized for tasks where computational resources are limited. It retains much of the performance of its larger counterpart but is designed to run efficiently on smaller devices.
\item \textbf{Mistral-7B}~\cite{jiang2023mistral} is a dense transformer model with 7 billion parameters, developed by Mistral AI. It offers strong language understanding and generation capabilities while being smaller in scale compared to other large models, balancing efficiency and performance.
\item \textbf{OPT-13B}~\cite{abs220501068} is a model developed by Meta, designed to be a large-scale generative language model. With 13 billion parameters, it is optimized for large-scale language tasks and is designed to provide open access to competitive performance, fostering transparency in AI development.
\end{itemize}

\textbf{Fine-Tuning strategies}
\begin{itemize}[leftmargin=*]
\item \textbf{BitFit}~\cite{zaken2022bitfit} is a lightweight parameter-efficient fine-tuning method. Unlike traditional fine-tuning approaches that modify all model parameters, BitFit focuses only on fine-tuning the bias terms of the model. This significantly reduces the computational cost and memory requirements while still achieving competitive performance on downstream tasks. It is particularly useful when fine-tuning large models on smaller datasets.
\item \textbf{AdaLoRA}~\cite{zhang2023adaptive} is an adaptive parameter-efficient fine-tuning method. It dynamically adjusts the rank of low-rank matrices during fine-tuning, allowing for efficient adaptation to specific tasks. AdaLoRA achieves a balance between efficiency and accuracy by selecting optimal ranks, enabling models to generalize well with minimal computational overhead.
\item \textbf{Prefix}~\cite{lester2021power} is a fine-tuning method that adds a small number of task-specific parameters (prefix tokens) to the input of the model. This method allows the pre-trained model to adapt to different tasks by simply learning the right combination of these prefix tokens, making it a lightweight and computationally efficient approach for transfer learning.
\item \textbf{LoRA}~\cite{hu2021lora} is an efficient fine-tuning method for large-scale pre-trained language models. Its core idea is to reduce the amount of trainable parameters in the fine-tuning process through low-rank matrix decomposition while maintaining model performance. It is proposed mainly to solve the high cost of traditional full-parameter fine-tuning in terms of computing resources, memory usage and deployment costs.
\item \textbf{IA3}~\cite{liu2022few} is a parameter efficient fine-tuning method that aims to achieve performance comparable to full parameter fine-tuning of large models with minimal parameter adjustments. The core idea is to dynamically scale the internal activations of the model through learning vectors, so as to adapt to downstream tasks without significantly increasing computing resources.
\item \textbf{FFT} refers to updating all model parameters for downstream tasks based on pre-trained models to adapt them to specific task requirements. Although parameter efficient fine-tuning methods have become increasingly popular in recent years, FFT is still the most powerful method for obtaining the strongest baseline in many scenarios.
\end{itemize}

\section{Datasets}
\label{app:datasets}
\paragraph{Adversarial benchmarks} include \textbf{Adversarial SQuAD} and \textbf{AdvGLUE}.
\textbf{Adversarial SQuAD} is an adversarial question answering dataset, primarily consisting of several sub-datasets:
\begin{itemize}[leftmargin=*]
\item \textbf{AddSent (AS)}~\cite{JiaL17}: A grammatical adversarial test set with 1k instances, in which misleading texts are generated from questions through rules and crowdsourcing.

\item \textbf{AddAny (AA)}~\cite{JiaL17}: Ungrammatical adversarial test set with 1k instances, in which misleading texts are automatically generated according to question words and common words.

\item \textbf{AddAnyExtend (AAE)}~\cite{liu2020robust}: Extended AddAny with 2.6k instances, which contains not only question words but also high-frequency words, passage words, and random common words.

\item \textbf{AddAnsCtx (AAC)}~\cite{liu2020robust}: Answer context test set with 10k instances, where the misleading texts are answer sentences with the answer tokens removed.

\item \textbf{AddNegAns (ANA)}~\cite{liu2020robust}: Negative expression test set with 5k instances, which uses negative expressions of fake answers as misleading texts.
\end{itemize}
\textbf{Adversarial GLUE}~\cite{WangXWG0GA021} is a multi-task benchmark for robustness evaluation, mainly including \textbf{SST2}, \textbf{QQP}, \textbf{MNLI}, \textbf{MNLI-MM}, \textbf{QNLI} and \textbf{RTE}.
\paragraph{Performance datasets}
\begin{itemize}[leftmargin=*]

\item \textbf{ViGGO}~\cite{juraska2019viggo} is an English dataset for data-to-text generation, focusing on video game opinions. The original task involves transforming a functional representation (a collection of attribute-values) into coherent text that includes those attributes. In this paper, we will reverse this process: converting unstructured text into a structured and parsable functional representation. This representation encapsulates the information found in the text and can be used for indexing and other downstream applications.
\item \textbf{GSM8k}~\cite{cobbe2021training} is a standard academic benchmark for evaluating language models on math reasoning and understanding. The goal of this dataset is to test the reasoning ability of language models.
\item \textbf{SQL Generation}~\cite{zhongSeq2SQL2017,yu2018spider} is a combination of the WikiSQL~\cite{zhongSeq2SQL2017} and Spider~\cite{yu2018spider}. This dataset contains 78,577 examples. Each example consists of a natural language query, corresponding SQL CREATE TABLE statements, and the SQL query corresponding to the natural language question. The goal is to take the natural language query and SQL CREATE TABLE statements as context and produce a SQL output that can query the given SQL tables and produce an output that answers the natural language query.
\item \textbf{SQUAD}~\cite{socher2013recursive} is a reading comprehension dataset consisting of questions posed by crowdworkers on a set of Wikipedia articles. SQuAD 1.1 contains over 100,000 question-answer pairs across 536 articles. It serves as a benchmark for evaluating machine reading comprehension systems. 
\item \textbf{SST-2}~\cite{socher2013recursive} is a sentiment analysis dataset derived from movie reviews. It contains 11,855 sentences, each labeled as either positive or negative. The dataset is parsed with Stanford's parser and includes 215,154 unique phrases, each annotated by three human judges.
\item \textbf{QQP}\footnote{https://quoradata.quora.com/First-Quora-Dataset-Release-Question-Pairs} consists of pairs of questions from Quora and is used for the task of identifying whether two questions are semantically equivalent. It has been widely used for training and evaluating models on paraphrase detection.
\item \textbf{MNLI}~\cite{williams2018broad} consists of sentence pairs labeled with one of three categories: entailment, contradiction, or neutral. It is designed to evaluate models on natural language inference (NLI) across different genres of text.
\item \textbf{MNLI-MM}~\cite{williams2018broad} includes sentence pairs that are more "matched" in terms of domain and linguistic features, providing a more controlled environment for NLI evaluation.
\item \textbf{QNLI}~\cite{rajpurkar2018know} is a binary classification task where the goal is to determine if a given passage contains the answer to a question.
\item \textbf{RTE}~\cite{wang2018glue} contains sentence pairs labeled as either entailment or non-entailment. It is used for evaluating models on the task of recognizing whether a hypothesis sentence can be logically inferred from a premise sentence.
\item \textbf{EmoC}~\cite{2019-semeval} focuses on contextual emotion detection in dialogues. It consists of English tweets annotated with four emotion classes: happy, sad, angry, and others. The dataset includes three-turn dialogues to capture emotional nuances in conversational contexts.
\item \textbf{TREC}~\cite{voorhees2000building} is designed for question classification tasks. It contains 5,500 labeled questions in the training set and 500 in the test set, categorized into six coarse classes and 50 fine classes.
\item \textbf{Amazon}~\cite{mcauley2013hidden} comprises product reviews and metadata from Amazon, including ratings, text, helpfulness votes, descriptions, prices, and images. It spans from May 1996 to July 2014, offering a rich resource for sentiment analysis and recommendation system research. 
\item \textbf{AGNews}~\cite{zhang2015character} is a news classification dataset containing 127,600 training samples and 7,600 test samples across four categories: World, Sports, Business, and Sci/Tech. It is constructed from the AG's corpus of news articles and is commonly used for evaluating text classification models.
\item \textbf{MT-bench}~\cite{zheng2024judging} is a benchmark designed to evaluate the fine-grained abilities of large language models (LLMs) in multi-turn dialogues. It includes 3,388 expert-level pairwise human preferences for model responses to 80 questions, assessing aspects like reasoning, coding, and math. Please refer to MT-bench~\cite{zheng2024judging} for more detailed information. Note that we fine-tune language models on Alpaca~\cite{taori2023stanford}, which consists of 51K instruction-following examples generated using text-davinci-003 (GPT-3.5).

\end{itemize}

\section{Implementation Details}
\label{app:implementation_details}
The performance results of the experiment are based on training with the AdamW optimizer. We selected the optimal hyperparameters, such as batch size and learning rate, through a grid search. For few-shot learning, we take one demonstration per class to form the prompt\footnote{The templates of prompts are presented in \Cref{app:prompt}} and append the sample to be predicted at the end of the prompt. We experiment with different numbers of training samples: 5, 10, 20, 50, 100, and 200 samples per class. All results are averaged over five random seeds (i.e., 0, 42, 421, 520, and 1228).

For AdvGLUE and Adversarial SQuAD, we train models and select the best model based on the original training and validation sets (i.e., GLUE and SQuAD), and then use the selected model to test on the adversarial test set. Since the test set of Adversarial SQuAD is not open source, we use the adversarial validation set as the test set. We report the key hyperparameter configurations for all experiments in Table~\ref{tab:hift_hyper} (\Cref{app:hyper}).

\section{Prompts}
\label{app:prompt}
We implement prompt-based fine-tuning for few-shot learning tasks.
The goal is to predict the correct class given a few examples. We reformulate the task as a language modeling problem. Let $M$ be a language model with vocabulary $V$, and let $\mathcal{L}$ be a set of label words. The training set $\mathcal{T}$ consists of pairs $(s, l)$, where
$s$ is a sequence of tokens from the vocabulary $V$ and $l$ is a label word from the set $\mathcal{L}$. In a sentiment analysis task, for instance, we define a pattern $\mathcal{P}(s,l)$ which associates a text $s=$`Nice performance' and a label word $l=$`Positive' as follows:
$$
\colorbox{LightSteelBlue1}{\text{Review: \underline{Nice performance.} Sentiment: \underline{Positive}}}
$$
For a $k$-class classification task, we sample one demonstration per class from the training set $\mathcal{T}$, and concatenate them with the text $s$ to be classified to form the prompt $X(s)$:
\begin{equation}
\label{equa_prompt}
X(s) = \mathcal{P}(s_1,l_1)\oplus \ldots \oplus \mathcal{P}(s_k,l_k)\oplus \mathcal{P}(s,\varepsilon)
\end{equation}
$\oplus$ denotes the concatenation of the input demonstrations and $\varepsilon$ is the empty string.

\section{Limitations}
\label{sec:limitations}
We do not provide a strict definition of task complexity, we empirically observe that tasks such as reasoning are generally more complex than classification tasks. We focus solely on a single PEFT method and do not extend our investigation to hybrid PEFT approaches. Our robustness evaluation is conducted from the user's perspective and is limited to the model's reasoning performance, without accounting for training-phase attack methods such as gradient-based attacks.

\section{Proofs}

\subsection{Proof of Non-surjective Function}
\label{proof:1}
\begin{proof}[Proof of \Cref{theorem:1}]

Given a continuously differentiable map $g$: $\mathbb{R}^k \rightarrow \mathbb{R}^d$, where $k \ll d$, then $g$ is a non-surjective function.

Define the set of all possible points that can be mapped from $\mathbb{R}^k$ to $\mathbb{R}^d$ via $g$:
\begin{equation}
Im(g) = \{g(\Phi) \in \mathbb{R}^d | \Phi \in \mathbb{R}^k\}
\end{equation}
When $g$ is a full rank mapping, we can have:
\begin{equation}
Im(g) = g(\mathbb{R}^k) \subset \mathbb{R}^d
\end{equation}
The image $Im(g)$ is a $k$-dimensional embedded submanifold of an $d$-dimensional space. According to Sard’s theorem\cite{sard1942measure}, when $k \ll d$, $Im(g)$, viewed as a subset of $\mathbb{R}^d$, has Lebesgue measure zero in $\mathbb{R}^d$. This shows that the $g$ cannot cover the $\mathbb{R}^d$ and is a non-surjective mapping. This conclusion is further strengthened when $g$ is a non-full rank mapping.
\end{proof}

\subsection{PEFT Capacity Upper Bound}
\label{proof:2}

\begin{proof}[Proof of \Cref{theorem:2}]
For input $x \in D$, we have:
\begin{equation}
\label{equation:6}
f(x) = \sigma((W_0 + \Phi)x)
\end{equation}
where $\sigma$ is the activation function that typically satisfies the L-Lipschitz condition. For any $u,v \in \mathbb{R}^d, \|\sigma(u) -v\| \le L\|u-v\|$ always holds, where L is the Lipschitz constant of the activation function $\sigma$. The incremental difference after fine-tuning is:
\begin{align}
\label{equation:7}
\|f(x) - f_0(x)\| = \|\sigma(W_0x + \Phi x) - \sigma(W_0x)\| {\leq} L\|\Phi x\| \leq L \|\Phi\| \|x\|
\end{align}
So for a certain layer we have:
\begin{equation}
\label{equation:8}
\forall x, \|f(x) - f_0(x)\| \leq L \|\Phi\| \|x\|
\end{equation}
Extending this theory to a network of depth $N$, the original parameters of each layer, along with the parameters introduced by PEFT, are denoted as follows:
\begin{equation}
\label{equation:9}
\begin{aligned}
&f_0(x) = \sigma_N(W_{0,N}\sigma_{N-1}(\cdots\sigma_1(W_{0,1}x))),\\
& f(x) = \sigma_N((W_{0,N} + \Phi_N)\sigma_{N-1}(\cdots\sigma_1((W_{0,1} + \Phi_1)x)))
\end{aligned}
\end{equation}
The activation Lipschitz constant of the kth layer is $L_k$, and the input-output norm is consistent with $\|\|$. Let $x_0^k=\sigma_k(W_{0,k}x_0^{k-1})$, $x^k=\sigma_k(W_0^k+\Phi_k)x^{k-1}$, where $x_0^k$ denotes the original representation of the foundation model at layer $k$, and $x^k$ denotes the representation of the fine-tuned model at layer $k$. The difference in representation between the fine-tuned and foundation models is:
\begin{equation}
\label{equation:10}
\begin{aligned}
\|x^{k} - x_0^{k}\| &= \|\sigma_k(W_{0,k}x^{(k-1)} + \Phi_k x^{(k-1)}) - \sigma_k(W_{0,k}x_0^{(k-1)})\| \\
&\leq L_k \|W_{0,k}x^{(k-1)} + \Phi_k x^{(k-1)} - W_{0,k}x_0^{(k-1)}\| \\
&\leq L_k \left(\|W_{0,k}\| \|x^{(k-1)} - x_0^{(k-1)}\| + \|\Phi_k\| \|x^{(k-1)}\|\right)
\end{aligned}
\end{equation}
Using recursion we can get the overall output difference:
\begin{equation}
\label{equation:11}
\|f(x) - f_0(x)\| \leq \sum_{k=1}^{N} \left[ L_N\|W_{0,N}\| \cdots L_{k+1}\|W_{0,k+1}\| \times L_k\|\Phi_k\|\|x\| \right].
\end{equation}
Assume that the upper bounds of the weight norm and activation Lipschitz constant of each layer are $M$ and $L$ respectively, that is, $\|W_{0,k}\| \leq M, \quad L_k \leq L, \quad \forall k$. At the same time, let $\alpha_k \triangleq L \|\Phi_k\|$, we have:
\begin{equation}
\label{equation:12}
\|f(x) - f_0(x)\| \leq \sum_{k=1}^{N} M^{N-k} L^{N-k} \alpha_k \|x\|
\end{equation}
\end{proof}

\subsection{Rule of Diminishing Marginal Benefit}
\label{proof:3}
\begin{proof}[Proof of \Cref{theorem:3}]
In PEFT fine-tuning, there exists a critical parameter $\Phi_c$ such that when $Rank(\Phi) \geq r_c$, further increasing the parameter amount of $\Phi$ does not lead to significant improvements in the performance of the optimal solution to the optimization problem. 

For any given task, the ideal weight update $\Delta W^*$ can be expressed using singular value decomposition as follows:
\begin{equation}
\label{equation:13}
\Delta W^* = U \Sigma V^T = \sum_{i=1}^{r_c} \sigma_i u_i v_i^T
\end{equation}
where $\sigma_1 \geq \sigma_2 \geq \dots \geq \sigma_{r_c} \geq 0$ are the singular values, and $u_i$ and $v_i$ are the corresponding left and right singular vectors, respectively. According to the Eckart–Young–Mirsky theorem~\cite{Mirsky1960SYMMETRICGF}, the best rank $r$ approximation is given by:
\begin{equation}
\label{equation:14}
\Delta W_r = \sum_{i=1}^r \sigma_i u_i v_i^T
\end{equation}
Under the Frobenius norm, the approximation error is:
\begin{equation}
\label{equation:15}
\left\| \Delta W^* - \Delta W_r \right\|_F^2 = \sum_{i=r+1}^{r_c} \sigma_i^2
\end{equation}
Existing studies have shown that singular values typically decay exponentially\cite{Feng2022RankDI}; therefore, we assume $\sigma_i \approx O(e^{-\lambda i})$, where $\lambda > 0$. accordingly, the critical rank $r_c$ satisfies the following:
\begin{equation}
\label{equation:16}
\sum_{i=r_c+1}^{rank(\Phi)} \sigma_i^2 < \epsilon
\end{equation}
Where $\epsilon$ is a positive number related to the generalization error. Since the loss function is typically smooth and continuous, there exists a constant $L > 0$ such that:
\begin{equation}
\label{equation:17}
|\mathcal{L} (W + \Delta W_1) - \mathcal{L}(W + \Delta W_2)| \le L\|\Delta W_1-\Delta W_2\|_F
\end{equation}
Therefore, when $r\ge r_c$, we have:
\begin{equation}
\label{equation:18}
|\mathcal{L} (W + \Delta W^*) - \mathcal{L}(W + \Delta W_r)| \le L\|\Delta W^*-\Delta W_r\|_F \le L \sqrt(\epsilon)
\end{equation}
Statistical learning theory shows that over-parameterized models can lead to: 
\begin{equation}
\label{equation:19}
\mathcal{L}_{\mathcal{D}_{test}}(W + \Delta W_r) - \mathcal{L}_{\mathcal{D}_{train}}(W + \Delta W_r) \propto \sqrt{\frac{r}{N}}
\end{equation}
Where $N$ is the number of training samples. Based on the above analysis, when $r \ge r_c$, the following conclusions can be drawn:
\begin{equation}
\label{equation:20}
\mathcal{L}_{\mathcal{D}_{test}}(W + \Delta W_r) \leq \mathcal{L}_{\mathcal{D}_{test}}(W + \Delta W^*) + L\sqrt{\epsilon} + O(\sqrt{\frac{r}{N}})
\end{equation}
As $r$ continues to increase, the first term remains unchanged, the second term approaches zero, but the third term (generalization error) increases, leading to a plateau or even a decline in overall performance.
\end{proof}

\subsection{PEFT More Sensitive to Disturbances}
\label{proof:4}
\begin{proof}[Proof of \Cref{theorem:4}]
Given the perturbed data distribution $D'$, 
\begin{equation}
\label{equation:21}
\theta^{\prime}_{perturbed} = \text{arg min}_{\theta \in \Theta} \mathbb{E}_{(x,y)\thicksim D'}[\mathcal{L}(f(x;\theta),y)]
\end{equation}
The optimization variable in PEFT is $\Phi$, in contrast to FFT, where it is $\theta$. Define the optimization variable $\delta \in \{\Phi,\theta\}$. By Taylor expanding the loss function, we can have:
\begin{equation}
\label{equation:22}
\mathcal{L}(\delta + \Delta \delta) \approx \mathcal{L}(\delta) + \nabla_\delta\mathcal{L}(\delta)^T\Delta \delta + \frac{1}{2}\Delta \delta^T H\Delta \delta
\end{equation}
where $H=\nabla_\delta^2\mathcal{L}(\delta)$ is Hessian Matrix. Assume the disturbance factor is $\epsilon$, the loss objective is:
\begin{equation}
\label{equation:23}
\mathcal{L}(\delta + \Delta \delta) \approx \mathcal{L}(\delta) + (\nabla_\delta\mathcal{L}(\delta)^T+\epsilon)\Delta \delta + \frac{1}{2}\Delta \delta^T H\Delta \delta
\end{equation}
Determine the extreme value of the quadratic function with respect to $\Delta \delta$:
\begin{equation}
\label{equation:24}
\nabla_{\Delta\delta} \left[ (\nabla_\delta\mathcal{L} + \epsilon)^T \Delta\delta + \frac{1}{2}\Delta\delta^T H\Delta\delta \right] = 0
\end{equation}
Taking the derivative, we get:
\begin{equation}
\label{equation:25}
\nabla_{\delta}\mathcal{L} + \epsilon + H\Delta\delta = 0 \Rightarrow \Delta \delta = -H^{-1}(\nabla_\delta\mathcal{L} + \epsilon)
\end{equation}
For FFT, $H$ is a positive definite or semi-definite matrix and invertible, we can have:
\begin{equation}
\label{equation:26}
\Delta \theta_{full} = -H^{-1}(\nabla_\theta\mathcal{L} + \epsilon)
\end{equation}
Substituting $\Delta \theta_{full}$ into the Taylor expansion of the loss function, and noting that $\Delta \theta_{full}$ is a first-order optimal solution—causing the first-order term to vanish—we obtain:
\begin{equation}
\label{equation:27}
\Delta\mathcal{L}_{\text{full}} \approx \frac{1}{2}\Delta\theta^T H\Delta\theta
\end{equation}
Substitute $\Delta \theta = -H^{-1}(\nabla_\theta \mathcal{L} + \epsilon)$:
\begin{equation}
\begin{aligned}
\label{equation:28}
\Delta\mathcal{L}_{\text{full}} &\approx \frac{1}{2} \left[-H^{-1}(\nabla_\theta\mathcal{L} + \epsilon)\right]^T H \left[-H^{-1}(\nabla_\theta\mathcal{L} + \epsilon)\right] \\
&= \frac{1}{2}(\nabla_\theta\mathcal{L} + \epsilon)^T H^{-1} H H^{-1} (\nabla_\theta\mathcal{L} + \epsilon) \\
&= \frac{1}{2}(\nabla_\theta\mathcal{L} + \epsilon)^T H^{-1} (\nabla_\theta\mathcal{L} + \epsilon)
\end{aligned}
\end{equation}
We only focus on the disturbance term, then:
\begin{equation}
\label{equation:29}
\Delta \mathcal{L}_{full} \approx \frac{1}{2}\epsilon^TH^{-1}\epsilon
\end{equation}

For PEFT, $H$ is typically a $d \times k$ matrix with $k \ll d$, indicating that $H$ is non-invertible (i.e., not exist $H^{-1}$). According to Theorem~\ref{theorem:1}, we can represent the parameter update as a projection mapping from $\mathbb{R}^k$ to $\mathbb{R}^d$:
\begin{equation}
\label{equation:30}
\theta = \theta_0 + g(\Phi)
\end{equation}
Here we view $g$ as a linear layer structure, and let $g=P\Phi$. we have:
\begin{equation}
\nabla_{\Phi}\mathcal{L} = \frac{\partial\mathcal{L}}{\partial\theta} \cdot \frac{\partial\theta}{\partial\Phi} = \nabla_{\theta}\mathcal{L} \cdot P = P^T\nabla_{\theta}\mathcal{L}
\end{equation}
The further Hessian matrix $H_{\Phi}$ is:
\begin{equation}
H_{\Phi} = \nabla^2_{\Phi}\mathcal{L} = \frac{\partial^2 \mathcal{L}}{\partial\Phi^2}=\left(\frac{\partial\theta}{\partial\Phi}\right)^T \cdot \frac{\partial^2 \mathcal{L}}{\partial\theta^2} \cdot \left(\frac{\partial\theta}{\partial\Phi}\right) = P^THP
\end{equation}
According Equation~\ref{equation:11}, the parameter update of PEFT is:
\begin{equation}
\Delta\theta_{peft} = -PH_{\Phi}^{+}P^T(\nabla_\theta\mathcal{L} + \epsilon)
\end{equation}
where $H_{\Phi}^+$ is the pseudo-inverse of $P^THP$ (because it may not be full rank). Since PEFT operates within a restricted low-dimensional space, the perturbation in $\mathbb{R}^d$ can be expressed as:
\begin{equation}
\epsilon = \epsilon_{\parallel} + \epsilon_{\perp}
\end{equation}
where $\epsilon_{\parallel}$ is the projection of the perturbation in the PEFT subspace, and $\epsilon_{\perp}$ is the component orthogonal to this subspace. PEFT can only counteract the component of the perturbation within the subspace, $\epsilon_{\parallel} = PP^T\epsilon$, and cannot counteract the perpendicular component $\epsilon_{\perp} = (I-PP^T)\epsilon$. The loss variation of PEFT induced by the perturbation is given by:
\begin{equation}
\Delta \mathcal{L}_{peft} \approx \frac{1}{2}\epsilon_{\parallel}^TH_{\Phi}^{+}\epsilon_{\parallel} + \epsilon_{\perp}^T \nabla_\theta\mathcal{L}
\end{equation}
Let $\Delta\mathcal{L}_{\text{peft}} - \Delta\mathcal{L}_{\text{full}}$ have:
\begin{equation}
\Delta\mathcal{L}_{\text{peft}} - \Delta\mathcal{L}_{\text{full}} = \frac{1}{2}\epsilon_{\parallel}^TH_{\Phi}^+\epsilon_{\parallel} + \epsilon_{\perp}^T\nabla_{\theta}\mathcal{L} - \frac{1}{2}\epsilon^TH^{-1}\epsilon
\end{equation}
Expanding $\epsilon = \epsilon_{\parallel} + \epsilon_{\perp}$ yields:
\begin{equation}
\epsilon^T H^{-1} \epsilon = (\epsilon_{\parallel} + \epsilon_{\perp})^T H^{-1}(\epsilon_{\parallel} + \epsilon_{\perp}) = \epsilon_{\parallel}^T H^{-1} \epsilon_{\parallel} + \epsilon_{\perp}^T H^{-1} \epsilon_{\perp} + 2\epsilon_{\parallel}^T H^{-1} \epsilon_{\perp}
\end{equation}
Since $\epsilon_{\parallel} \perp \epsilon_{\perp}$, they belong to orthogonal subspaces, then:
\begin{equation}
\epsilon_{\parallel}^TH^{-1}\epsilon_{\perp}=0
\end{equation}
We have:
\begin{equation}
\Delta\mathcal{L}_{\text{peft}} - \Delta\mathcal{L}_{\text{full}} = \frac{1}{2}(\epsilon_{\parallel}^TH_{\Phi}^+\epsilon_{\parallel} - \epsilon_{\parallel}^TH^{-1}\epsilon_{\parallel}) + \epsilon_{\perp}^T\nabla_\theta\mathcal{L} - \frac{1}{2}\epsilon_{\perp}^TH^{-1}\epsilon_{\perp}
\end{equation}
PEFT fine-tunes a model within a low-dimensional subspace, which can be interpreted as an projection from $\mathbb{R}^d$ to $\mathbb{R}^k$. Therefore:
\begin{equation}
\epsilon_{\parallel}^TH_{\Phi}^+\epsilon_{\parallel} \geq \epsilon_{\parallel}^TH^{-1}\epsilon_{\parallel}
\end{equation}
The equality $=$ holds \textit{if and only if} the subspace spanned by $P$ coincides with the subspace containing $\epsilon_{\parallel}$. Consider the term $\epsilon_{\perp}^T\nabla_\theta\mathcal{L} - \frac{1}{2}\epsilon_{\perp}^TH^{-1}\epsilon_{\perp}$, we have:
\begin{equation}
\begin{aligned}
&\epsilon_{\perp}^T\nabla_\theta\mathcal{L} - \frac{1}{2}\epsilon_{\perp}^TH^{-1}\epsilon_{\perp} = \frac{1}{2} \left[2\epsilon_{\perp}^T\nabla_\theta\mathcal{L}-\epsilon_{\perp}^TH^{-1}\epsilon_{\perp}\right] \\
&=\frac{1}{2}[\nabla_\theta\mathcal{L}^T H\nabla_\theta\mathcal{L} - (\epsilon_{\perp} - H\nabla_\theta\mathcal{L})^T H^{-1}(\epsilon_{\perp} - H\nabla_\theta\mathcal{L})]
\end{aligned}
\end{equation}
For convenience, $\pi$ is used to denote $\nabla_\theta\mathcal{L}$, and $x$ to denote $\epsilon_{\perp}$.
Expanding the quadratic term, we can have:
\begin{equation}
(x - H\pi)^T H^{-1}(x - H\pi) = x^T H^{-1}x - 2\pi^T H^Tx + \pi^T H^TA\pi = x^T H^{-1}x - 2\pi^T Hx + \pi^TH\pi
\end{equation}
The whole equation becomes:
\begin{equation}
\frac{1}{2}\left[\pi^T H\pi - x^T H^{-1}x + 2\pi^T Hx - \pi^T H\pi\right] = \frac{1}{2}\left[2\pi^T Hx - x^T H^{-1}x\right]
\end{equation}
For convenience, we let $y=H^{1/2}x$, $x=H^{-1/2}y$, the equation becomes $\pi^TH^{1/2}y-\frac{1}{2}y^TH^{-2}y$. This formula is a convex quadratic function of $y$, and its maximum occurs at:
\begin{equation}
\nabla_y = H^{1/2}\pi - H^{-2}y = 0 \Rightarrow y^* = H^{5/2}\pi
\end{equation}
Then we can have:
\begin{equation}
\begin{aligned}
&\pi^T H^{1/2}y^* = \pi^T H^{1/2}H^{5/2}\pi = \pi^T H^3 \pi\\
&(y^*)^T H^{-2}y^* = \pi^T H^{5/2}H^{-2}H^{5/2}\pi = \pi^T Hg
\end{aligned}
\end{equation}
Finally we have:
\begin{equation}
\pi^T Hx - \frac{1}{2}x^T H^{-1}x = \pi^T H^3\pi - \frac{1}{2}\pi^T H\pi > 0
\end{equation}
Since $H \succ 0$, it follows that $H^3 \succ H$, which means:
\begin{equation}
\pi^T H^3g > \pi^T H\pi \Rightarrow \pi^T H^3\pi - \frac{1}{2}\pi^T H\pi > 0
\end{equation}
Finally we have:
\begin{equation}
\epsilon_{\perp}^T\nabla_\theta\mathcal{L} - \frac{1}{2}\epsilon_{\perp}^TH^{-1}\epsilon_{\perp} =
\frac{1}{2}\left[\nabla_\theta\mathcal{L}^T H\nabla_\theta\mathcal{L} - (\epsilon_{\perp} - H\nabla_\theta\mathcal{L})^T H^{-1}(\epsilon_{\perp} - H\nabla_\theta\mathcal{L})\right] > 0
\end{equation}
That is to say, $\Delta\mathcal{L}_{\text{peft}} - \Delta\mathcal{L}_{\text{full}}> 0$.
\end{proof}

\section{Hyperparameter}
\label{app:hyper}
For LLaMa2 family models on ViGGO, GSM8k and SQL generation tasks, please use the scripts provided in the link: \url{https://github.com/ray-project/ray/tree/master/doc/source/templates/04_finetuning_llms_with_deepspeed}. The parameter configuration of LoRA is as follows: rank is 8,
Lora\_alpha is 16,
Target\_modules list is \{ "q\_proj", "v\_proj", "k\_proj", "o\_proj", "gate\_proj", "up\_proj", "down\_proj", "embed\_tokens", "lm\_head" \}.
For prefix-tuning, the virtual vocabulary is set to 20 and the rank of AdaLoRA is 8, lora\_alpha is 32. 

\begin{table*}[h]
\centering
\small
\caption{The hyperparameter grids used for HiFT experiments.}
\label{tab:hift_hyper}
\begin{tabular}{lrc}
\toprule
Experiment & Hyperparameters & Values \\
\midrule
RoBERTa-base & Total Batch size & $64$ \\
& Learning rate & $\{1\mathrm{e}{-5}, 2\mathrm{e}{-5}, 3\mathrm{e}{-5} \}$ \\
& warmup & \{0.0, 0.02, 0.06\} \\
& Device & 8*GTX 1080Ti (11G) \\
& Weight Decay & $0$ \\
\midrule
RoBERTa-large & Total Batch size & $32$ \\
& Learning rate & $\{1\mathrm{e}{-5}, 2\mathrm{e}{-5}, 3\mathrm{e}{-5} \}$ \\
& warmup & \{0.0, 0.02, 0.06\} \\
& Device & 8*GTX 1080Ti (11G) \\
& Weight Decay & $0$ \\
\midrule
OPT-13B & Batch size & $\{2,4,8\}$ \\
& Learning Rates & $\{1\mathrm{e}{-5}, 2\mathrm{e}{-5}, 5\mathrm{e}{-5},8\mathrm{e}{-5}\}$ \\
& Device & A100 (80G) \\
& Weight Decay & $0$\\
\midrule
Mistral-7B & Batch size & $\{2,4,8\}$ \\
& Learning Rates & $\{1\mathrm{e}{-5}, 2\mathrm{e}{-5}, 5\mathrm{e}{-5}\}$ \\
& Device & A100 (80G) \\
& Weight Decay & $0$\\
\midrule
TinyLLaMA & Batch size & $\{2,4,8\}$ \\
& Learning Rates & $\{2\mathrm{e}{-5}, 5\mathrm{e}{-5},8\mathrm{e}{-5}\}$ \\
& Device & A100 (80G) \\
& Weight Decay & $0$\\
\midrule
LLaMA2-7B & Batch size & $\{2,4,8\}$ \\
& Learning Rates & $\{1\mathrm{e}{-5}, 2\mathrm{e}{-5}, 5\mathrm{e}{-5},8\mathrm{e}{-5}\}$ \\
& Device & A100 (80G) \\
& Weight Decay & $0$\\
\midrule
LLaMA2-13B & Batch size & $\{2,4,8\}$ \\
& Learning Rates & $\{1\mathrm{e}{-5}, 2\mathrm{e}{-5}, 5\mathrm{e}{-5},8\mathrm{e}{-5}\}$ \\
& Device & A100 (80G) \\
& Weight Decay & $0$\\
\midrule
LLaMA2-70B & Batch size & $\{2,4,8\}$ \\
& Learning Rates & $\{1\mathrm{e}{-5}, 2\mathrm{e}{-5}, 5\mathrm{e}{-5},8\mathrm{e}{-5}\}$ \\
& Device & A100 (80G) \\
& Weight Decay & $0$\\
\bottomrule
\end{tabular}
\end{table*}

\section{More Experiments}
\label{sec:more_experiments}
\begin{table*}[!t]
\centering
\caption{The impact of training samples on the performance of different fine-tuning methods. $k$ denotes the number of training examples per class. The base model is LLaMA2-7B.}
\label{tab:llama2-few}
\begin{adjustbox}{max width=0.98\textwidth}
\includegraphics[width=\textwidth]{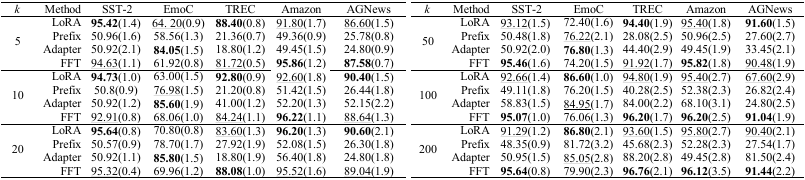}
\end{adjustbox}
\end{table*}

\begin{table*}[!t]
\caption{The effect of the amount of fine-tuning parameters on model performance under standard fine-tuning on AdvGLUE.}
\label{tab:abla_glue}
\begin{adjustbox}{max width=\textwidth, center}
\includegraphics[width=\textwidth]{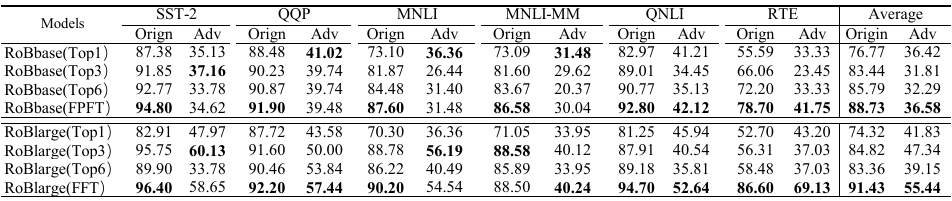}
\end{adjustbox}
\end{table*}

\begin{table*}[!t]
\caption{The effect of the amount of fine-tuning parameters on model performance under standard fine-tuning on Adversarial SQuAD dataset.}
\label{tab:abla_squad}
\begin{adjustbox}{max width=\textwidth, center}
\includegraphics[width=\textwidth]{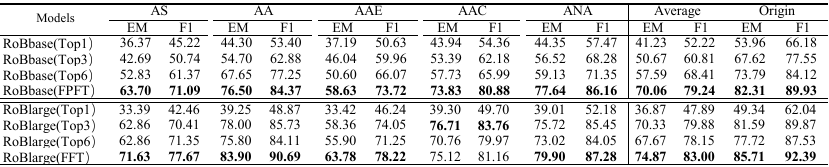}
\end{adjustbox}
\end{table*}

\end{document}